\documentclass{article} % For LaTeX2e
\usepackage{iclr2025,times}

\usepackage{amsmath,amsfonts,bm}

\def\eqref#1{equation~\ref{#1}}
\def\1{\bm{1}}

\def\mX{{\bm{X}}}

\DeclareMathAlphabet{\mathsfit}{\encodingdefault}{\sfdefault}{m}{sl}
\SetMathAlphabet{\mathsfit}{bold}{\encodingdefault}{\sfdefault}{bx}{n}

\def\gG{{\mathcal{G}}}

\usepackage{hyperref}
\usepackage{url}
\usepackage{xcolor}

\title{On the Limits of Applying Graph Transformers for Brain Connectome Classification}

\author{Jose Lara-Rangel, Clare Heinbaugh \thanks{Equal contribution.} \\
Department of Engineering\\
University of Cambridge\\
Cambridge, UK \\
\texttt{\{jml224,ceh210\}@cam.ac.uk} \\
}

\usepackage{graphicx} 
\usepackage{subcaption}

\iclrfinalcopy % Uncomment for camera-ready version, but NOT for submission.
\begin{document}

\maketitle

\begin{abstract}
Brain connectomes offer detailed maps of neural connections within the brain. Recent studies have proposed novel connectome graph datasets and attempted to improve connectome classification by using graph deep learning. With recent advances demonstrating transformers' ability to model intricate relationships and outperform in various domains, this work explores their performance on the novel NeuroGraph benchmark datasets and synthetic variants derived from probabilistically removing edges to simulate noisy data. Our findings suggest that graph transformers offer no major advantage over traditional GNNs on this dataset. Furthermore, both traditional and transformer GNN models maintain accuracy even with all edges removed, suggesting that the dataset's graph structures may not significantly impact predictions. We propose further assessing NeuroGraph as a brain connectome benchmark, emphasizing the need for well-curated datasets and improved preprocessing strategies to obtain meaningful edge connections.

\end{abstract}

\section{Introduction}

The human brain is a complex network of interconnected regions. Neuroscientists often divide the brain into Regions of Interest (ROIs) and measure signaling between these regions called connectomes ~\citep{NeuroGraph_cit}. Functional magnetic resonance imaging (fMRI) can be used to observe signals between different regions. Like other areas of biology, the question of how to preprocess the data is often as crucial as the modelling step \citep{preprocessing_bio_example}. Typically, mapping features signal measurements to graph structures requires domain-specific knowledge of neuroscience. To automate the workflow, ~\citet{NeuroGraph_cit} recently presented NeuroGraph as a collection of benchmark graph datasets derived from the Human Connectome Project ~\citep{hcp_cit}, giving graph machine learning researchers access to brain connectome graph data.

After testing a variety of deep learning and graph machine learning models on NeuroGraph, ~\citet{NeuroGraph_cit} identified the ResidualGCN (or GNN*), which uses graph convolutions and residual connections, as the top performer. However, it exhibited overfitting and poor generalization \citep{NeuroGraph_cit}. Following the trend in deep learning to move towards transformers and attention-based models \citep{attention_is_all_you_need, ISLAM2024122666, shehzad2024graph}, we explore the application of Exphormer—a novel sparse Graph Transformer (GT)—along with other attention mechanisms for brain connectome gender, activity, and age prediction tasks, comparing their performance to previous models applied on the NeuroGraph benchmark \citep{exphormer_cit, NeuroGraph_cit}. Surprisingly, while attention was expected to improve performance and generalization \citep{hussain_global_sa}, Exphormer’s performance matched that of ResidualGCN without surpassing it. Despite extensive regularization efforts, Exphormer also showed signs of overfitting and did not improve performance on the NeuroGraph datasets.

We decided to investigate other potential advantages of GTs and whether the comparable performance with previous models in \cite{NeuroGraph_cit}, including
graph and non-graph deep learning models, could be attributed to the models themselves or the graph structures in the NeuroGraph datasets. To this end, we generated synthetic datasets by removing edges, an approach that has been previously explored for other purposes \citep{han2025rethinkingfunctionalbrainconnectome}. We hypothesized that given Exphormer’s ability to model local and global relationships, combined with its structural encodings, it would be more resilient to edge-dropping compared to ResidualGCN, which relies heavily on local neighborhoods \citep{exphormer_cit, hussain_global_sa}. Surprisingly, both models showed resilience, maintaining performance even with complete edge removal. This suggests that node relationships may not be critical for predictions, with models potentially relying solely on node features. Consequently, the graph structure in the NeuroGraph benchmark datasets require further evaluation.

Our contributions are: 
\begin{itemize} 
    \item We explore the application of sparse GTs for brain connectome classification tasks, showing the potential of GTs in understanding complex neural connections. 
    \item We generate synthetic datasets with missing edges and show that even with significant alterations to graph structures both ResidualGCN and Exphormer preserve performance compared to the orignal datasets.
    \item We propose further evaluation of the NeuroGraph static classification datasets as benchmarks, highlighting their potential simplicity and susceptibility to overfitting, and advocate for alternative preprocessing strategies to create more meaningful graph structures. 
\end{itemize}

\section*{Background}

\subsection*{NeuroGraph Dataset}

NeuroGraph was introduced in \citet{NeuroGraph_cit} as a comprehensive collection of static and dynamic brain connectome datasets tailored for graph machine learning classification and regression tasks. We focus on classification tasks on demographics and activity state static graphs, due to data accessibility, computational simplicity, and existing model performance comparisons. The demographic group includes gender classification as male and female, and age classification into 22-25, 26-30, and 31-35 years groups. The activity state task is to identify core functions relevant to brain-cognition-behavior relationships. This consists of seven distinct activities: emotional processing, gambling, language, motor, relational processing, social cognition, and working memory. Hence, the NeuroGraph datasets considered are HCP-Activity, HCP-Gender, and HCP-Age. 

In a static graph representation, individual data points are represented by single graphs. In the static connectome graph, $\displaystyle \gG = (V, E, \mX)$, the node set $V$ represents ROIs, and the edge set $E\subseteq V \times V$ the positive correlations between pairs of ROIs determined with a defined threshold, and considers a feature matrix $\displaystyle \mX \in \mathbb{R}^{n \times d}$, where $n$ is the total number of ROIs and $d$ is the dimension of the feature vector. In NeuroGraph, the correlation vectors are used as node features \citep{NeuroGraph_cit}.

\subsection*{Sparse Graph Transformers}

GTs extend conventional transformers for graph representation learning ~\citep{graph_transformer, attention_is_all_you_need, min2022transformer, dwivedi2020generalization}. Unlike traditional Message-Passing Graph Neural Networks (MPGNNs), GTs operate on a fully connected graph to better model long-range dependencies and interactions among distant nodes. In contrast to local attention in Graph Attention Networks (GATs), GTs allow every node to communicate directly with all others, producing a global attention mechanism ~\citep{veliko}. By incorporating the input graph's structure as a soft inductive bias and retaining the global attention mechanism of traditional transformers, GTs effectively leverage global graph structural information. Consequently, GTs mitigate common shortcomings observed in MPGNNs, such as over-smoothing, over-squashing, and limited expressivity \citep{graph_transformer, min2022transformer}. 

However, in the dense GT setting, where the graph is fully connected, the computational complexity becomes $O(|V|^2)$ \citep{min2022transformer, dwivedi2020generalization}. Real-world graph datasets, like the NeuroGraph case, often exhibit arbitrary connectivity structures shaped by the application domain, sometimes with an extremely large number of nodes. Although this diversity can offer valuable information, it also makes the use of GTs impractical. Sparse GTs address this issue by restricting attention to specific node pairs, reducing complexity while preserving critical structural information.

Exphormer is a sparse GT introduced in ~\citet{exphormer_cit} that integrates two sparse attention mechanisms: global nodes and expander graphs. Global nodes are connected to all other nodes, whereas the number of edges in the expander graph is proportional to the number of nodes. For an input graph $\displaystyle \gG = (V,E)$, Exphormer builds an undirected interaction graph with three components: \textbf{expander graph attention} for long-range connections using \( O(|V|) \) edges, \textbf{global attention} with virtual nodes linked to all others for universal representation, and \textbf{local neighborhood attention} linking immediate neighbors with \( O(|E|) \) edges. Expander graphs act as sparse approximations of complete graphs, retaining the spectral properties of a full attention mechanism while requiring $O(|V|)$ edges. Unlike dense attention mechanisms, Exphormer's sparse attention mechanisms do not directly connect all pairs of nodes in the pairwise interaction graph, but by stacking transformer layers most nodes' pairwise interactions can be effectively modeled \citep{exphormer_cit}. Thus, Exphormer has an expander-based sparse attention mechanism with $O(|V| + |E|)$ computational complexity that achieved results comparable to a full transformer ~\citep{exphormer_cit}.

\vspace{-0.5em}
\section*{Methodology}
\vspace{-0.1em}

To evaluate Exphormer performance, we consider ResidualGCN as a baseline ~\citep{NeuroGraph_cit}. ResidualGCN's residual nature is derived from the concatenation of outputs from a series of graph convolutional layers in a sequential model, passed into a final multi-layer perceptron. We also explored the introduction of attention to this base architecture. However, it did not produce improvements in performance or other notable benefits, see Appendix \ref{subsec:additional_exphormer}. 

For the Exphormer, we experimented with different numbers of layers, dropout rates for the network and attention mechanism, and numbers of attention heads. The final configuration used dropout probability of $0.1$, attention dropout of $0.3$, $2$ layers, and $4$ attention heads. All experiments used learning rate decay starting at $0.001$, decaying by $1e-5$, over a total of $100$ epochs with $5$ warmup epochs. We used three different seeds for both the Exphormer and ResidualGCN and assessed the alignment with the results in \citet{NeuroGraph_cit}, which only included one run for each experiment.

Apart from evaluating performance, we investigated potential advantages of using attention-based models. Our hypothesis was that the attention mechanism could enhance robustness to data noise, particularly in scenarios where certain graph structure components, such as edges, are missing. To verify that the graph structure, nodes and edges taken together, convey meaningful information for prediction, it is important to compare models under noisy or incomplete data settings. We simulate noisy incomplete data by removing edges based on a pre-specified probability of edge removal.

\vspace{-0.5em}
\section*{Experiments \& discussion}

\subsection*{performance on NeuroGraph datasets}

Table \ref{tab:dropped-edges} shows the performance of both models for each dataset. During training, Exphormer achieves its best performance in the early stages of the training process, whereas ResidualGCN reaches its best performance towards the end of training, see Appendix \ref{subsec:additional_exphormer}. However, the sparse transformer mechanism of Exphormer does not provide clear performance improvements compared to ResidualGCN. Both models performed similarly across datasets, with ResidualGCN slightly outperforming Exphormer on the HCP-Activity dataset and solidly outperforming on the HCP-Gender dataset. Although the validation and test curves remained similar in most experiments, both models overfit the data. HCP-Age exhibits the most prominent performance drop of approximately 50\% between train and test accuracy, see Appendix \ref{subsec:additional_exphormer} for additional details.

Regularizing with dropout applied to both the network and attention mechanism, along with reducing layers, did not significantly improve performance. Fewer layers sped up training, higher network dropout slowed learning and reduced accuracy, while attention dropout stabilized learning. No configuration outperformed the baseline, see Appendix \ref{subsec:additional_exphormer}.

\subsection*{Dropped edges synthetic dataset}

Table \ref{tab:dropped-edges} shows the results of the edge-dropping experiments. As expected, Exphormer maintained performance as the edge-drop probability increased. In contrast, and contrary to what we anticipated, the performance of ResidualGCN also remained unchanged within error margins as edge-drop probabilities increased, even in the extreme case where all edges were removed.

Since removing edges had no substantial impact on the models' performance, and our prediction tasks focus on graph-level properties rather than node classification, it is possible that the node features in this dataset alone provide sufficient information to make predictions. Alternatively, the edges in the NeuroGraph datasets may carry irrelevant information that adds little to the prediction tasks, or the specific graph structure itself may not be critical for these tasks. This could also explain the lack of improvement with Exphormer, as some of its characteristics rely on edge features. Interestingly, ResidualGCN appears to stabilize as edges are removed while Exphormer fluctuates, especially considering the margins of error for the HCP-Age dataset in \ref{tab:dropped-edges}. While ResidualGCN and Exphormer capture local neighborhood attention, Exphormer also uses global attention, thus, future work could examine how global attention is better leveraged by datasets beyond NeuroGraph.

Having tested models with varying complexities that yielded similar results across the learning process, we can attribute this performance consistency without edges to factors beyond the network architecture’s capacity, see Appendix \ref{subsec:additional_attention}. This could also explain why a vanilla neural network achieved competitive performance with the ResidualGCN in \cite{NeuroGraph_cit}. This aligns with the recent work of \citet{learningHigherOrderRelationshipsOfBrainRegions} suggesting that pair-wise relationships (graph edges) may be insufficient in the application context of brain connectivity and warrants capturing higher-order relationships between ROIs. Consequently, further improvement would be expected from better pre-processing strategies rather than from applying attention-based models to current graph structures.

Unlike recent studies such as \citet{han2025rethinkingfunctionalbrainconnectome} that focused on model architectures, our experiments relate the lack of improvement to the graph structures in the NeuroGraph datasets. By experimenting with attention-based mechanisms, we circumvent the claim of \citet{han2025rethinkingfunctionalbrainconnectome} that complex graph theory models, like MPGNNs, exhibit poor performance compared to simpler models due to over-squashing. Instead, we hypothesize that the stagnant performance of attention-based graph deep learning models compared to traditional deep learning approaches could be explained by the lack of edge-conveyed information. Thus, neuroscience expertise is still necessary to pre-process brain connectomes to create meaningful graph benchmark datasets.

\begin{table}[t]
\caption{Test accuracy for models trained on NeuroGraph datasets with varying probabilities of dropped edges. Average results over 3 seeds.}
\vspace{-0.2em}
\label{tab:dropped-edges}
\begin{center}
\begin{tabular}{cccc}
\multicolumn{1}{c}{\bf Dataset} & \multicolumn{1}{c}{\bf Drop edge probability} & \multicolumn{1}{c}{\bf ResidualGCN} & \multicolumn{1}{c}{\bf Exphormer} \\
\hline \\

HCP-Activity & 0.0 & $98.10 \pm 0.20$ & $97.54 \pm 0.20$ \\
             & 0.5 & $97.85 \pm 0.14$ & $97.40 \pm 0.69$ \\
             & 1.0 & $97.76 \pm 0.08$ & $97.63 \pm 0.47$ \\
\hline \\

HCP-Gender   & 0.0 & $88.27 \pm 1.16$ & $80.87 \pm 3.25$ \\
             & 0.5 & $84.26 \pm 1.61$ & $79.94 \pm 1.41$ \\
             & 1.0 & $86.58 \pm 2.02$ & $79.01 \pm 0.54$ \\
\hline \\

HCP-Age      & 0.0 & $50.23 \pm 2.35$ & $48.91 \pm 4.21$ \\
             & 0.5 & $50.39 \pm 0.71$ & $49.53 \pm 2.81$ \\
             & 1.0 & $52.11 \pm 0.00$ & $47.35 \pm 5.15$ \\

\end{tabular}
\end{center}
\vspace{-1.4em}
\end{table}

\vspace{-0.5em}
\section*{Conclusion \& future work}
Contrary to expectations, applying sparse graph transformers like Exphormer, and other attention-based models, to static connectome classification tasks did not improve performance. Although Exphormer showed faster convergence, it performed worse compared to ResidualGCN, originally proposed in \citet{NeuroGraph_cit}, and exhibited overfitting despite regularization efforts. In experiments simulating noisy conditions through random edge removal, both models' performance remained unaffected, even without edges. These findings highlight the need to further investigate the relevance of the graph structures in the NeuroGraph benchmark datasets and alternative fMRI processing pipelines to generate meaningful graph representations. This study highlights the importance of well-curated data in complex tasks like brain connectome classification.

\bibliography{iclr2025}
\bibliographystyle{iclr2025}

\newpage
\appendix 
\section*{Appendix. Additional experimental results}

\subsection*{Additional Exphormer results}
\label{subsec:additional_exphormer}

\begin{figure}[htbp]
  \centering
  \begin{subfigure}[b]{0.45\textwidth}
    \centering
    \includegraphics[width=\textwidth]{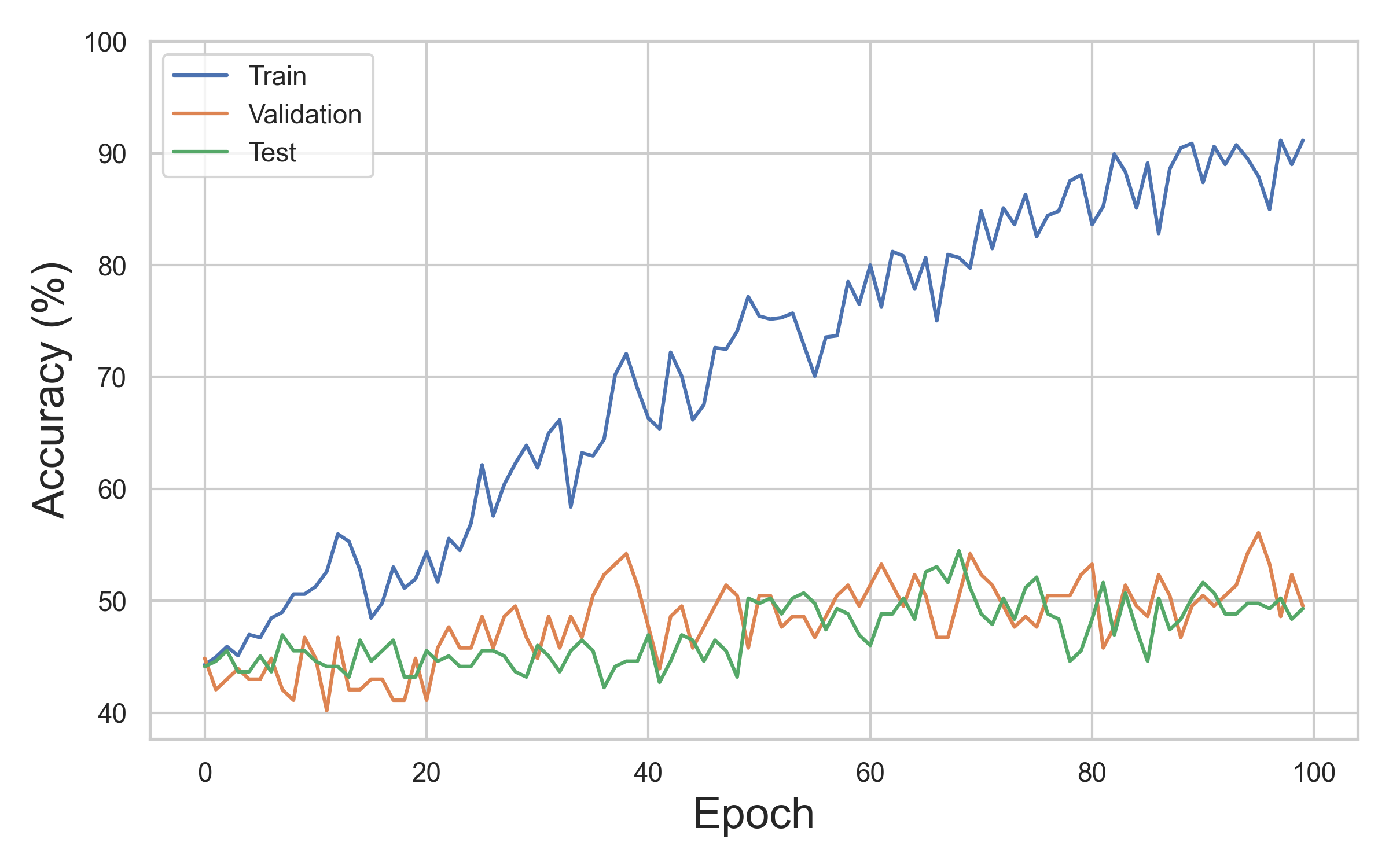}
    \caption{Accuracy vs. epoch number for ResidualGCN trained on the original HCP-Age dataset.}
    \label{fig:plot1}
  \end{subfigure}
  \hfill
  \begin{subfigure}[b]{0.45\textwidth}
    \centering
    \includegraphics[width=\textwidth]{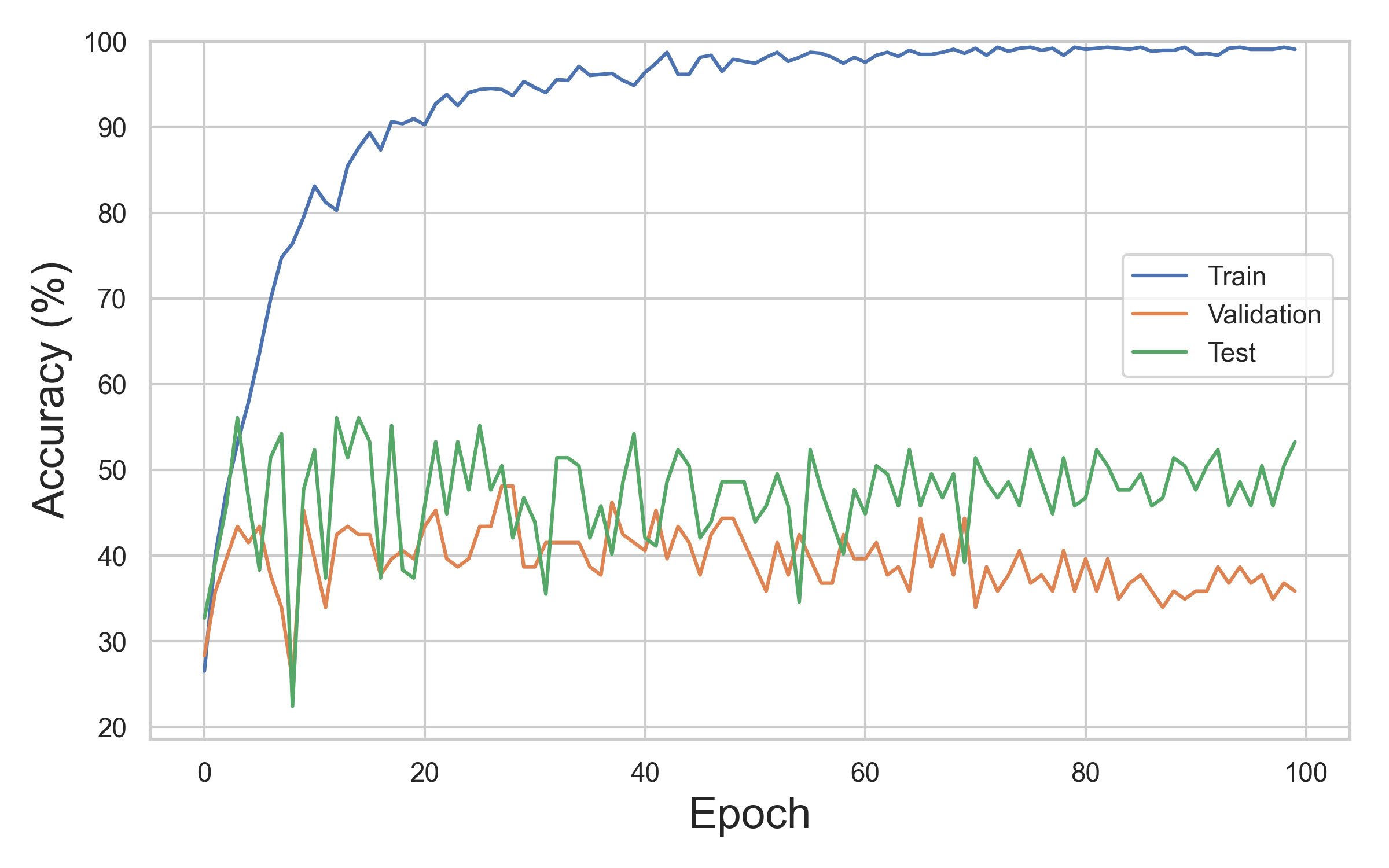}
    \caption{Accuracy vs. epoch number for Exphormer trained on the original HCP-Age dataset.}
    \label{fig:plot2}
  \end{subfigure}
  \caption{Accuracy curves for both models on HPC-Age.}
    \label{fig:accuracy_curves}
\end{figure}

Figure \ref{fig:accuracy_curves} shows the learning curves for both the ResidualGCN and Exphormer on the HPC-Age datasets, illustrating that both models reach comparable accuracies but exhibit clear overfitting. Among the tasks and datasets considered, HCP-Age displays the largest drop between training and test accuracies, and the sparse GT properties of Exphormer \citep{exphormer_cit} did not yield significant improvements over the baseline. However, it is worth noting that Exphormer converges to its optimal performance within fewer epochs, whereas ResidualGCN typically requires almost the entire training schedule to reach its best performance.

Similar trends are observed in other HPC datasets. Validation and test curves remain similar, with Exphormer converging faster than ResidualGCN, but even when the gap between training and testing is less pronounced, there remains a distinct sign of overfitting in most experiments, and neither model decisively outperforms the other.

Trying to alleviate overfitting, we applied dropout to both the network layers and the Exphormer attention mechanism, varying dropout probabilities from 0.1 to 0.6. The best results were achieved with combinations of dropout probabilities of 0.1 and 0.3. The results for HCP-Gender and HCP-Activity are shown in Tables \ref{tab:attention_dropout} and \ref{tab:attention_dropout_activity}, respectively. Increasing the network dropout slowed learning and generally reduced accuracy, whereas moderate attention dropout helped stabilize the learning curves. Nevertheless, no configuration outperformed the baseline by a substantial margin on any of the datasets. This suggests that while dropout can help control overfitting, these particular tasks may require additional strategies to improve performance.

\begin{table}[htbp]
\caption{Validation accuracy for HCP-Gender with different combinations of dropout.}
\label{tab:attention_dropout}
    \begin{center}
    \begin{tabular}{cccc}
    \multicolumn{2}{c}{\bf Gender} & \multicolumn{2}{c}{\bf Attention Dropout} \\
    %\hline \\
    \multicolumn{2}{c}{} & 0.1 & 0.3 \\
    \hline \\
    \bf Dropout & 0.1 & 84.26 & 85.18 \\
    \bf         & 0.3 & 80.56 & 82.41 \\
    \end{tabular}
    \end{center}
\end{table}

\begin{table}[htbp]
\caption{Test accuracy for Activity with different combinations of dropout.}
\label{tab:attention_dropout_activity}
    \begin{center}
    \begin{tabular}{ccccc}
    \multicolumn{2}{c}{\bf Activity} & \multicolumn{3}{c}{\bf Attention Dropout} \\
    %\hline \\
    \multicolumn{2}{c}{} & 0.1 & 0.3 & 0.5 \\
    \hline \\
    \bf Dropout & 0.1 & 96.24 & 96.94 & 96.91 \\
    \bf         & 0.3 & 96.64 & 96.78 & 95.97 \\
    \end{tabular}
    \end{center}
\end{table}

We also explored using different numbers of layers. As shown in Table \ref{tab:different_layers}, 2-layer and 3-layer configurations achieved similar performance, although the 2-layer model trained significantly faster. A 5-layer version was also tested but proved to be much more time-consuming without providing noticeable gains, so we discontinued further experiments.

\begin{table}[htbp]
\caption{Validation and test accuracy for Exphormer experiments with 2 and 3 layers.}
\label{tab:different_layers}
    \begin{center}
    \begin{tabular}{ccccccc}
    \multicolumn{1}{c}{\bf Layers} & \multicolumn{2}{c}{\bf Gender} & \multicolumn{2}{c}{\bf Age} & \multicolumn{2}{c}{\bf Activity} \\
    %\hline \\
    & \bf Val & \bf Test & \bf Val & \bf Test & \bf Val & \bf Test \\
    \hline \\
    2 layers & 84.26 & 77.78 & 54.72 & 48.60 & 97.45 & 97.36 \\
    3 layers & 84.26 & 79.63 & 52.83 & 49.53 & 97.71 & 96.78 \\
    \end{tabular}
    \end{center}
\end{table}

\subsection*{Additional attention results}
\label{subsec:additional_attention}

Exphormer builds upon GraphGPS \citep{rampavsek2022recipe}, a modular framework that combines local message passing with a global attention mechanism and leverages various positional and structural encodings. This design provides a flexible platform for incorporating Transformer layers alongside MPNN models. As reported by \cite{exphormer_cit}, some experiments showed improved Exphormer performance when additional GraphGPS layers were included. Specifically, by replacing the Transformer component in GraphGPS with an Exphormer layer, Exphormer achieved accuracies comparable to a full Transformer-based GraphGPS. Moreover, when Exphormer was combined with MPNNs in GraphGPS, it reached State-of-the-Art (SOTA) or near-SOTA results \citep{exphormer_cit, rampavsek2022recipe}. Motivated by these findings, we tested the combination of GraphGPS and Exphormer on the NeuroGraph static datasets. However, as shown in Table \ref{tab:alternative_model_scores} for the case of HCP-Gender, this hybrid setup did not yield improvements.

Furthermore, we extended our analysis to assess whether introducing attention layers in ResidualGCN could provide improvements in predictive accuracy. We compared two alternative attention mechanisms different to the one used in \cite{NeuroGraph_cit}. The first one consists of an attention layer after each GCN layer, while the second version includes an attention layer after the concatenation of the GCN layers and the residual connections. Moreover, we extended the experiment to apply the attention mechanism according to a specified probability; with a probability of 1, it always applies attention. None of these modifications improved performance. Table \ref{tab:alternative_model_scores} exemplifies the validation and test accuracies obtained on HCP-Gender for these alternatives. In some cases, the models with added attention matched or slightly exceeded the baseline accuracy but did not establish a clear advantage overall. Despite the different attention placements and probabilities, none of these modifications consistently improved performance.

\begin{table}[htbp]
\caption{Validation and test accuracy for alternative models.}
\label{tab:alternative_model_scores}
    \begin{center}
    \begin{tabular}{lcc}
    \multicolumn{1}{c}{\bf Model} & \multicolumn{1}{c}{\bf Validation} & \multicolumn{1}{c}{\bf Test} \\
    \hline \\
    
    GPS+Exphormer & 82.40 & 80.55 \\
    ResidualGCN+Attention (1) & 87.25 & 88.25 \\
    ResidualGCN+Attention (0.8) & 86.25 & 87.00 \\
    ResidualGCN+Attention (0.3) & 84.75 & 88.25 \\
    ResidualGCN+Attention2 (1) & 83.50 & 87.00 \\
    ResidualGCN+Attention2 (0.6) & 79.50 & 85.00 \\
    
    \end{tabular}
    \end{center}
\end{table}

\subsection*{Additional details on NeuroGraph}
\label{subsec:additional_neurograph}

There are only a few options for brain connectome datasets \citep{NeuroGraph_cit_27, NeuroGraph_cit_3, NeuroGraph_cit,
learningHigherOrderRelationshipsOfBrainRegions}, which often require specialized processing to be used for graph machine learning. In this work, we explore NeuroGraph, a recent collection of static and dynamic brain connectome benchmark datasets tailored for graph machine learning classification and regression tasks introduced in \citet{NeuroGraph_cit}. NeuroGraph was built using publicly available datasets derived from the Human Connectome Project (HCP), applying minimal preprocessing steps based on standard fMRI procedures to create static and dynamic graph representations ~\citep{hcp_cit}. In a static graph representation, individual data points are represented by single graphs. Conversely, dynamic graph representations encapsulate time-varying interactions and connectivity patterns within the brain. In our study, we focused on classification tasks on static graphs, due to dataset accessibility, computational simplicity, and existing model performance comparisons.

The static connectome graphs in NeuroGraph comprise node sets representing ROIs and edge sets representing positive correlations between pairs of ROIs. These benchmark datasets are categorized into three distinct groups: demographics, activity states, and cognitive traits. We focus on the first two groups, considering HCP-Activity, HCP-Gender, and HCP-Age. Dataset statistics are provided in Table \ref{tab:datasets}.

\begin{table}[htbp]
\caption{Dataset information for HCP-Activity, HCP-Gender, and HCP-Age \citep{NeuroGraph_cit}.}
\label{tab:datasets}
    \begin{center}
    \begin{tabular}{lcccc}
    \multicolumn{1}{c}{\bf Dataset} & \multicolumn{1}{c}{\bf \# of graphs} & \multicolumn{1}{c}{\bf \# of node features} & \multicolumn{1}{c}{\bf \# of edge features} & \multicolumn{1}{c}{\bf \# of classes} \\
    \hline \\
    
    HCP-Activity & 7332 & 400 & 1 & 7 \\
    HCP-Gender & 1078 & 1000 & 1 & 2 \\
    HCP-Age & 1065 & 1000 & 1 & 3 \\
    
    \end{tabular}
    \end{center}
\end{table}

\end{document}